\documentclass[lettersize,journal]{IEEEtran}
\usepackage{amsmath,amsfonts}
\usepackage{algorithmic}
\usepackage{algorithm}
\usepackage{array}
\usepackage[caption=false,font=footnotesize,labelfont=rm,textfont=rm]{subfig}
\usepackage{textcomp}
\usepackage{stfloats}
\usepackage{url}
\usepackage{verbatim}
\usepackage{graphicx}
\usepackage{cite}
\usepackage{graphicx}

\usepackage{hyperref}  

\usepackage{cleveref}

\usepackage{amsmath}  

\usepackage{amsthm,amsmath,amssymb}

\usepackage{mathrsfs}

\usepackage{algorithmic}
\usepackage{algorithm}

\usepackage{tabularray}
\usepackage{color}

\usepackage{pifont}  
\usepackage{amssymb} 

\usepackage{cleveref}
\crefname{figure}{Figure}{Figures}
\Crefname{figure}{Figure}{Figures}

\crefname{figure}{Fig.}{Figs.}
\Crefname{figure}{Fig.}{Figs.}
\begin{document}

\title{Beamforming and Resource Allocation for Delay Minimization in RIS-Assisted OFDM Systems}

\author{
	\IEEEauthorblockN{
		Yu Ma,
		Xiao Li, 
		\IEEEmembership{Member, IEEE},
		Chongtao Guo,
		\IEEEmembership{Member, IEEE}, 
		Le Liang, 
		\IEEEmembership{Member, IEEE}, \\
		Michail Matthaiou,
		\IEEEmembership{Fellow, IEEE},
		and Shi Jin, 
		\IEEEmembership{Fellow, IEEE}
	}
	
	\thanks{Y. Ma, X. Li, and S. Jin are with the National Mobile Communications Research Laboratory, Southeast University, Nanjing 210096, China (e-mail: yuma@seu.edu.cn; li\_xiao@seu.edu.cn; jinshi@seu.edu.cn).
		
		C. Guo is with the College of Electronics and Information Engineering, Shenzhen University, Shenzhen 518060, China (e-mail: ctguo@szu.edu.cn).
		
		L. Liang is with the National Mobile Communications Research Laboratory, Southeast University, Nanjing 210096, China, and also with the Purple Mountain Laboratories, Nanjing 211111, China (e-mail: lliang@seu.edu.cn).
		
		M. Matthaiou is with the Centre for Wireless Innovation (CWI), Queen’s University Belfast, BT7 1NN Belfast, U.K. (e-mail: m.matthaiou@qub.ac.uk).
		}
}

\markboth{Journal of \LaTeX\ Class Files,~Vol.~14, No.~8, August~2021}%
{Shell \MakeLowercase{\textit{et al.}}: A Sample Article Using IEEEtran.cls for IEEE Journals}


\maketitle

\begin{abstract}

This paper investigates a joint beamforming and resource allocation problem in downlink reconfigurable intelligent surface (RIS)-assisted orthogonal frequency division multiplexing (OFDM) systems to minimize the average delay, where data packets for each user arrive at the base station (BS) stochastically. The sequential optimization problem is inherently a Markov decision process (MDP), thus falling within the remit of reinforcement learning. To effectively handle the mixed action space and reduce the state space dimensionality, a hybrid deep reinforcement learning (DRL) approach is proposed. Specifically, proximal policy optimization (PPO)-$\Theta$ is employed to optimize the RIS phase shift design, while PPO-$N$ is responsible for subcarrier allocation decisions. The active beamforming at the BS is then derived from the jointly optimized RIS phase shifts and subcarrier allocation decisions. To further mitigate the curse of dimensionality associated with subcarrier allocation, a multi-agent strategy is introduced to optimize the subcarrier allocation indicators more efficiently. Moreover, to achieve more adaptive resource allocation and accurately capture the network dynamics, key factors closely related to average delay—such as the number of backlogged packets in buffers and current packet arrivals are incorporated into the state space. Furthermore, a transfer learning framework is introduced to enhance the training efficiency and accelerate convergence. Simulation results demonstrate that the proposed algorithm significantly reduces the average delay, enhances the resource allocation efficiency, and achieves superior system robustness and fairness compared to baseline methods.

\end{abstract}


\section{Introduction}

\IEEEPARstart{W}{ith} the continuous evolution of wireless communication technologies, sixth-generation (6G) wireless communication systems are envisioned to achieve unprecedented performance in terms of capacity, latency, reliability, and energy efficiency\cite{6gg}. To meet the stringent demands of future applications, extensive research is being conducted on enhancing the fundamental capabilities of wireless systems~\cite{6g_thec}. Among the various enabling technologies, reconfigurable intelligent surfaces (RISs) have emerged as a key enabler\cite{enhancing_coverage}, drawing significant research interest due to their potential to reconfigure the wireless propagation environment \cite{sang22} and support enhanced system performance \cite{feng1, feng2}.

As a critical performance metric, latency plays a vital role in determining the overall performance of wireless systems, particularly for mission-critical applications, such as remote surgery, autonomous driving, factory automation, and intelligent transportation systems \cite{applications}. These applications often require real-time data exchange. In such scenarios, even a small increase in latency can lead to severe implications, such as system failures, road accidents, or disrupted services. Therefore, ensuring ultra-low latency, often on the order of tens of milliseconds or less \cite{delay_num}, is essential for numerous mission-critical applications in the 6G. 

However, given the inherently limited nature of wireless resources, ensuring low-latency communication in RIS-assisted wireless systems remains a fundamental challenge, where the phase design and resource allocation are coupled. In addition to the above mentioned challenge, limited spectrum  and power  resources constrain the system’s ability to support high data rates and real-time responsiveness, while multi-path effects introduce rapid fluctuations in the channel conditions and signal quality \cite{mutipath}, which necessiates network management to be adaptive to the environment. On the other hand, different users may have different numbers of backlogged packets, which needs to be taken into account in delay-aware resource allocation \cite{arq}. Therefore, in delay-aware RIS-assisted wireless networks, it is critical to make sequential resource allocation and phase design decisions while being adaptive to real-time channel conditions and queue statuses  \cite{9424177}, which is challenging due to complex network environments and mixed decision variables, including continuous RIS phase shift design and discrete variations of subcarrier allocation indicators.

\subsection{Prior Works and Motivation}
This section provides an overview of related work, with a summarized comparison shown in Table~\ref{tab:relatedwork}.
The delay refers to the duration from the instance the packet is generated at the transmitter to its reception by the receiver. This metric is crucial for evaluating the performance of communication networks, especially for applications requiring real-time responsiveness. Numerous studies (e.g., \cite{queuedelay1, queuedelay2, queuedelay3}) have shown that queuing delay constitutes the dominant component of the overall latency compared to other components, such as processing latency, transmission latency, and propagation latency.

To address delay-aware resource allocation, some works leverage large-scale fading for spectrum and power management. For instance, in device-to-device (D2D)-enabled vehicular networks, resource allocation based on large-scale fading improves vehicle-to-infrastructure (V2I) ergodic capacity, while ensuring vehicle-to-vehicle (V2V) reliability \cite{large-scale1}. Similarly, single-input multiple-output (SIMO) systems have adopted power and rate allocation strategies using average signal and interference statistics under reliability and energy efficiency constraints \cite{large-scale2}. Sun \emph{et al.} further proposed a resource block and power control algorithm for D2D-based V2V communications without relying on instantaneous channel state information (CSI) \cite{large-scale3}. However, these static methods struggle to adapt to rapid channel variations. To address this, Lyapunov optimization has gained traction for real-time delay-aware resource management, accounting for queue dynamics, channel fluctuations, and system constraints. For example, Guo \emph{et al.} developed a cross-layer Lyapunov framework combining transport-layer rate control and physical-layer allocation to meet delay targets \cite{Lyapunov}. Similar approaches have led to robust scheduling policies that ensure queue stability and service quality in dynamic environments \cite{Lyapunov1, Lyapunov2, Lyapunov3}.

\begin{table}[t]
	\centering
	\caption{Comparison of delay-oriented resource allocation methods}
	\label{tab:relatedwork}
	\resizebox{0.5\textwidth}{!}{ 
		\begin{tabular}{|l|c|c|c|c|c|}
			\hline
			\textbf{Research work} & \textbf{RIS} & \textbf{Delay-oriented} & \textbf{RL} & \textbf{Multi-agent} & \textbf{Queuing delay} \\
			\hline
			
			\cite{Lyapunov, Lyapunov1,Lyapunov2,Lyapunov3} & \ding{55} & \checkmark & \ding{55} & \ding{55} & \checkmark \\
			\cite{delay11}             & \ding{55} & \checkmark & \checkmark & \ding{55} & \checkmark \\
		
			\cite{RISDelay1}           & \checkmark & \checkmark  & \ding{55} & \ding{55} & \ding{55} \\
			\cite{RISDelay2}           & \checkmark & \checkmark  & \ding{55} & \ding{55} & \ding{55} \\
			\cite{RISLyapunov3}        & \checkmark & \checkmark & \ding{55} & \ding{55} & \checkmark  \\
			\textbf{The proposed work} & \checkmark & \checkmark & \checkmark & \checkmark & \checkmark \\
			\hline
		\end{tabular}
	}
\end{table}

Beyond conventional optimization techniques, deep reinforcement learning (DRL)-based algorithms have emerged as powerful tools for addressing delay-aware sequential resource allocation in wireless communication systems, offering a favorable balance between performance and computational complexity \cite{Federated_Learning}. Leveraging the Markov decision process (MDP) framework, DRL is particularly well-suited for real-time, sequential resource allocation decisions. Early efforts in this domain, such as the Q-learning-based link scheduling algorithm proposed in \cite{delay11}, demonstrate how queue length, channel conditions, and energy constraints can be jointly considered to minimize the average transmission delay. As the complexity of network environments increases, research has shifted from single-agent to multi-agent paradigms. For instance, multi-agent deep reinforcement learning (MADRL) has been adopted to better handle distributed decision-making scenarios. In \cite{8792382}, each V2V link was modeled as an autonomous agent that dynamically performs spectrum sharing and power control, effectively reducing latency while minimizing interference within V2I communications.

While RISs can theoretically alter the spectral efficiency, coverage, and energy efficiency \cite{Resource, Secure, Phase_Shifts, Energy_Efficiency, sang3}, their application in latency-sensitive scenarios remains limited. Most existing works focus on mobile edge computing and neglect queuing delay, which is crucial for real-time communication \cite{RISDelay1, RISDelay2, RISDelay3}. For instance, \cite{RISDelay1} minimized the task delay and energy in RIS-assisted non-orthogonal multiple access networks, while \cite{RISDelay2} addressed joint delay and rate in multiple-input multiple-output mobile edge computing systems.
To address this gap, recent studies adopt Lyapunov-based methods for queue-aware resource allocation, converting long-term delay goals into per-slot decisions to adapt to dynamic traffic \cite{RISLyapunov2, RISLyapunov3}. For example, \cite{RISLyapunov3} jointly optimized the power, channel assignment, and reflection coefficients in a semi-grant-free RIS-assisted system, improving the queue stability over RIS-free baselines.
However, current approaches remain insufficient. Many rely on complex alternating optimizations and fail to explicitly include queuing delay in their objective functions, limiting their effectiveness in truly delay-sensitive RIS-assisted systems.

Our research aims to bridge this gap by proposing a novel approach that integrates the RIS technology with advanced optimization methods, including DRL techniques, to minimize the average delay and enhance the network performance. Our approach provides a new perspective on addressing latency challenges in RIS-assisted systems and offers a fresh direction for improving the system performance in delay-sensitive applications. In general, this work seeks to unlock the untapped potential of RISs in managing queuing delay in practical communication systems, ultimately enhancing the efficiency of RIS-assisted wireless networks.

\subsection{Contributions and Organization}

This paper aims to investigate resource allocation methods for delay minimization in RIS-assisted networks by leveraging DRL, with a particular focus on maintaining user fairness. To achieve this goal, a hybrid DRL framework is proposed to tackle the challenge of delay optimization in such networks. The main contributions of this work are summarized as follows.
\begin{itemize}
	\item  A hybrid DRL framework is adopted, where the base station (BS) controller acts as the agent and is trained using proximal policy optimization (PPO). Specifically, PPO-$\Theta$ with continuous action space is used for the RIS phase shift optimization, while PPO-$N$ with discrete action space handles the subcarrier allocation. To address the curse of dimensionality, a multi-agent strategy is employed, where each agent is responsible for the allocation of a single subcarrier.
	
	\item To ensure low delay, fairness, and adaptability, the state space incorporates CSI, the number of arriving packets, and the number of backlogged packets in each buffer. This enables the agent to capture both channel conditions and delay status for more efficient resource allocation in dynamic environments.
	
	\item We leverage transfer learning to overcome slow convergence during the initial stage, which arises from weak incentives in the direct delay-aware reward design. Specifically, we first train the agent with the objective of maximizing the minimum rate.  Once a certain performance level is achieved, the pre-trained PPO-$\Theta$ is loaded as the initial model and further trained using the reward function and algorithm proposed in this work.
	
	\item We evaluate the proposed method across various RIS-assisted communication network configurations, demonstrating its superiority over baseline methods. Simulation results indicate that the method significantly reduces the average delay while ensuring user fairness, demonstrating its superiority over baseline methods. Specifically, fairness is enhanced in the balanced accumulation of data packets across user buffers, even under varying arrival rates. Moreover, the method exhibits excellent robustness, as it maintains stable performance under different delay taps, user distributions, and Ricean fading factors.

\end{itemize}

The rest of the paper is organized as follows: In Section \ref{section:System Model and Problem formulation}, the system model is introduced, followed by the formulation of the joint RIS phase and resource allocation optimization problem. At the end of this section, the optimization problem is formally defined. Section \ref{section:Deep Reinforcement Learning Based Algorithm} elaborates on  our propoosed hybrid DRL-based scheme, including the overall framework, PPO-$\Theta$ design, PPO-$N$ design, beamforming design, and trainning methodology. Section \ref{section:Numerical Results} presents the numerical simulations and complexity analysis to validate the proposed method. Finally, conclusions are presented in Section \ref{section:Conclusion}.

\begin{figure*}[!t]
	\centering
	
	\begin{minipage}[b]{\textwidth}
		\centering
		\subfloat[Illustration of a RIS-assisted OFDM-MISO system communication scenario]{\includegraphics[width=0.5\textwidth]{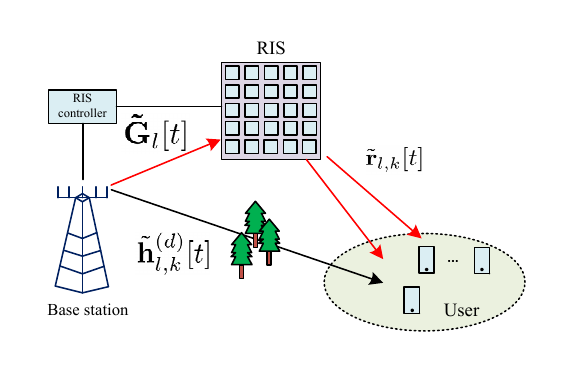}%
			\label{Fig:RIS}}
		\hfil
		\subfloat[Queueing model at the BS]{\includegraphics[width=0.5\textwidth]{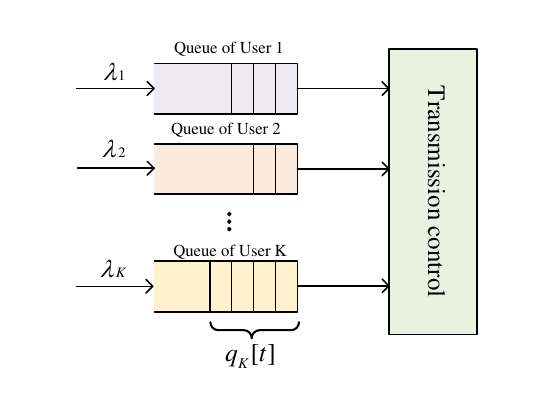}%
			\label{Fig:queue}}
		
	\end{minipage}

	\caption{System model.}
	\label{fig:model}
\end{figure*}

\subsection{Notations}
Scalar variables are denoted by normal-face letters $x$, while vectors and matrices are denoted by lower and upper case letters, $\mathbf{x}$ and $\mathbf{X}$, respectively. The sets of complex and real numbers are denoted by \(\mathbb{C}\) and \(\mathbb{R}\), respectively. The superscripts \((\cdot)^{T}\), \((\cdot)^{*}\), and \((\cdot)^{H}\) represent the transpose, conjugate, and conjugate transpose operations of a matrix, respectively. The notation \(\|\cdot\|\) represents the \(\ell_2\) norm of a vector. The expectation operator is denoted as \(\mathbb{E}\{\cdot\}\). The notation \(\text{diag}(\cdot)\) indicates a diagonal matrix construed by the entries of the input vector. The notation \(\mathcal{CN}(\mu, \sigma^2)\) is used to denote a circularly symmetric complex Gaussian distribution with mean $\mu$ and variance $\sigma^2$, while \((\cdot)^+\) refers to $\max(\cdot, 0)$.

\section{System Model and Problem formulation}   \label{section:System Model and Problem formulation}

We consider an RIS-assisted downlink multiple-input single-output (MISO) orthogonal frequency division multiplexing (OFDM) communication system, as illustrated in Fig. \ref{fig:model}\subref{Fig:RIS}, where a BS equipped with \( N_t \) antennas serves \( K \) single-antenna users. An RIS comprising \( M \) passive reflecting elements is deployed to effectively enhance the transmission quality. The entire communication duration is divided into \( T_{\mathrm{sum}} \) timeslots, indexed by \( \{1, 2, \ldots, t, \ldots, T_{\mathrm{sum}} \} \), each with a fixed length of \( T \). Within each time slot, the wireless channel is assumed to remain approximately static.
Each user's data packets, which arrive randomly at the corresponding buffer at the BS, follow a first-come, first-served (FCFS) discipline, as illustrated in Fig. \ref{fig:model}\subref{Fig:queue}. In each time slot, the accumulated packets in buffer \( k \) are scheduled for transmission to  user \( k \). Packets that cannot be delivered will remain in the buffer and continue to accumulate, resulting in increased packet delay over time.

\subsection{Signal Model}

In the MISO-OFDM system considered in this paper, $\tilde{\mathbf{h}}^{(d)}_{l,k}[t]\in\mathbb{C}^{1\times N_t},l=0,1,...,L_0-1$ represents the time-domain baseband equivalent channel of the direct link from the BS to user \( k \) at time slot $t$, where $L_0$ denotes the number of tap delays. Similarly, $\mathbf{\tilde{G}}_l[t]\in\mathbb{C}^{M\times N_t}, l=0,1,...,L_1 - 1$ and $\tilde{\mathbf{r}}_{l,k}[t]\in\mathbb{C}^{1\times M},l=0,1,...,L_2 - 1$ respectively denote the time-domain baseband equivalent channels from the BS to the RIS and from the RIS to  user $k$ at time slot $t$, where $L_1$ and $L_2$ are their respective numbers of delay taps. Therefore, the total maximum number of delay taps is $L=\operatorname*{max}\big\{L_{0},L_{1}+L_{2}-1\big\}$. Thus, the overall channel from the BS to user \( k \) at time slot $t$ can be expressed as
\begin{equation}
	\begin{aligned}
		\tilde{\mathbf{h}}_{l,k}[t]& =\tilde{\mathbf{h}}^{(d)}_{l,k}[t]+\sum_{i=0}^{L_{2}-1}\tilde{\mathbf{r}}_{i,k}[t]\boldsymbol{\Phi}[t]\tilde{\mathbf{G}}_{l-i}[t] \\
		&=\tilde{\mathbf{h}}^{(d)}_{l,k}[t]+\sum_{i=0}^{L_{2}-1}\boldsymbol{\mathbf{\varphi}}[t]^{T}\mathrm{diag}(\tilde{\mathbf{r}}_{i,k}[t])\tilde{\mathbf{G}}_{l-i}[t] \\
		&=\tilde{\mathbf{h}}^{(d)}_{l,k}[t]+\boldsymbol{\mathbf{\varphi}}[t]^{T}\sum_{i=0}^{L_{2}-1}\mathrm{diag}(\tilde{\mathbf{r}}_{i,k}[t])\tilde{\mathbf{G}}_{l-i}[t], \\ 
		& \quad \quad \quad  \quad l=0,1,...,L -1,
	\end{aligned}
\end{equation}
where $\tilde{\mathbf{G}}_{l}[t]=\mathbf{0} \text{,}\; l\in\begin{Bmatrix}1-L_{2}$\text{\,} $\ldots,-1\end{Bmatrix}\cup\begin{Bmatrix}L_{1},\ldots,L-1\end{Bmatrix}$, and $\boldsymbol{\Phi}[t]=\operatorname{diag}(e^{j\theta_{1}[t]},e^{j\theta_{2}[t]},...,e^{j\theta_{M}[t]})\in\mathbb{C}^{M\times M}$ is the  RIS reflection phase shift matrix at time slot $t$. Let $\boldsymbol{\mathbf{\varphi}}[t]=[e^{j\theta_{1}[t]},e^{j\theta_{2}[t]},...,e^{j\theta_{M}[t]}]^T\in\mathbb{C}^{M\times1}$ denote the RIS reflection phase shift vector at time slot $t$, where $\theta_{m}[t]$ refers to the phase shift of the reflection element $m$ at time slot $t$. We set $\tilde{\mathbf{H}}^{(r)}_{l,k}[t] = \sum_{i=0}^{L_{2}-1}\mathrm{diag}(\tilde{\mathbf{r}}_{i,k}[t])\tilde{\mathbf{G}}_{l-i}[t]$; thus, $\tilde{\mathbf{h}}_{l,k}[t]$ can be further expressed as

\begin{equation}
	\tilde{\mathbf{h}}_{l,k}[t]=\tilde{\mathbf{h}}^{(d)}_{l,k}[t]+\boldsymbol{\mathbf{\varphi}}[t]^{T}\tilde{\mathbf{H}}^{(r)}_{l,k}[t].
\end{equation}
To leverage the advantages of OFDM, we assume that the length of the cyclic prefix exceeds the maximum delay taps, i.e., \( N_{CP} \geq L \). This ensures that inter-symbol interference is eliminated. Furthermore, the discrete Fourier transform (DFT) is applied to transform the time-domain channel into the frequency-domain channel, which can be expressed as

\begin{equation}
	\begin{aligned}
		\bar{\mathbf{h}}_{n,k}[t]& =\sum_{l=0}^{L-1}\tilde{\mathbf{h}}_{l,k}[t] e^{\frac{-j2\pi ln}{N}} \\
		&=\sum_{l=0}^{L_{0}-1}\tilde{\mathbf{h}}^{(d)}_{l,k}[t] e^{\frac{-j2\pi ln}{N}}+\boldsymbol{\mathbf{\varphi}}[t]^{T}\sum_{l=0}^{L_{1}+L_{2}-2}\tilde{\mathbf{H}}^{(r)}_{l,k}[t] e^{\frac{-j2\pi ln}{N}} \\
		&=\bar{\mathbf{h}}^{(d)}_{n,k}[t]+{\boldsymbol{\varphi}}[t]^{T}\bar{\mathbf{H}}^{(r)}_{n,k}[t] ,
	\end{aligned}
\end{equation}
where $\bar{\mathbf{h}}^{(d)}_{n,k}[t]$ and $\bar{\mathbf{H}}^{(r)}_{n,k}[t]$ represent the direct link channel and cascaded reflection channel of user \( k \) on subcarrier \( n \)  at time  slot $t$ in the frequency domain, respectively. Therefore, the received signal of user \( k \) on subcarrier \( n \) at time  slot $t$ can be represented as

\begin{equation}
	y_{n,k}[t]=\big(\bar{\mathbf{h}}^{(d)}_{n,k}[t]+\boldsymbol{\varphi}^T\bar{\mathbf{H}}^{(r)}_{n,k}[t]\big)\mathbf{w}_{n}[t]s_{n}[t]+\nu_{n}[t], 
\end{equation}

\noindent where \( \mathbf{w}_{n}[t] \in \mathbb{C}^{N_t \times 1} \) denotes the beamforming vector on subcarrier \( n \) at time slot $t$, \( s_{n}[t] \) represents the transmitted signal on subcarrier \( n \)  at time slot $t$ with \( \mathbb{E}\{| s_{n}[t] |^2 \} = 1 \), while \( \nu_{n}[t] \) is additive Gaussian white noise, which satisfies \(\nu_{n}[t] \sim \mathcal{CN}(0, \sigma^2)\). Thus, the signal-to-noise ratio (SNR) of user \( k \) on subcarrier \( n \)  at time slot $t$ can be expressed as

\begin{equation}
	\gamma_{n,k}[t]=\frac{\left|(\bar{\mathbf{h}}^{(d)}_{n,k}[t]+\boldsymbol{\varphi}[t]^T\bar{\mathbf{H}}^{(r)}_{n,k}[t])\mathbf{w}_{n}[t]\right|^2}{\sigma^2}.
\end{equation}
Therefore, the achievable rate for user \( k \) at time slot $t$ is given by

\begin{equation}
	R_{k}[t]={W}\sum_{n=1}^{N}\alpha_{n,k}[t]\mathrm{log}_2(1+\gamma_{n,k}[t]),
\end{equation}
where \( W \) represents the bandwidth of the subcarrier, and \( \alpha_{n,k}[t] \) denotes the subcarrier allocation indicator at time slot $t$. In particular, we have $\alpha_{n,k}[t]=1$ if subcarrier $n$ is allocated to user $k$ at time slot $t$, and  $\alpha_{n,k}[t]=0$ otherwise. 

\subsection{Queueing Model}
In practice, packets arrive randomly at the transmitter and stored provisionally in a buffer before transmission to the receiver. Following the recommendation of the 3rd generation partnership project (3GPP), we model the packet arrival process as a Poisson process \cite{Passion}. Specifically, the number of packets arriving at buffer \( k \) during time slot \( t \), denoted by \( \ell_k[t] \), follows the probability density function

\begin{equation}
\mathrm {Pr}\{\ell_k[t]= i\}= \frac{(\lambda_k T)^i }{i!}e^{-\lambda_k T},
\end{equation}
where $\lambda_k$ is the average packet arrival rate of user $k$.  The number of packets that can be theoretically transmitted from the buffer \(k\) to  user \(k\) in time slot \(t\) is denoted as \(D_k[t]\)  given by
\begin{equation}
	D_k[t]=\left\lfloor\frac{TR_k\left[t\right]}{L}\right\rfloor,
\end{equation}
where $\lfloor \cdot \rfloor$ is the floor operation and $L$ is the number of bits contained in a packet.  
However, the number of packets that are eventually delivered during time slot $t$, denoted by $\Xi_k\left[t\right]$, is bounded by the number of accumulated packets in buffer $k$, as given by
\begin{equation}
	\Xi_k\left[t\right]=\min\left\{D_k\left[t\right],q_k\left[t\right]\right\},
\end{equation}
where $q_k\left[t\right]$ denotes the number of backlogged packets in buffer $k$ at the beginning of time slot $t$. Thus, the queue in the buffer $k$ is updated according to
\begin{equation}
	q_k\left[t+1\right]=q_k\left[t\right] + \ell_k\left[t\right] - \Xi_k\left[t\right].
\end{equation}

\subsection{Problem Formulation}
The delay experienced by each packet consists of two components: queueing delay and transmission delay, both of which are affected by the subcarrier allocation indicator, RIS reflection phase shifts, and the beamforming vector at the BS. Therefore, both delay components are considered in this paper. Let $m(k)$ represent the number of packets successfully delivered to user $k$ during the communication period with $T_{\text{sum}}$ time slots. The delay of the $g$-th packet of user $k$ during the considered time duration is denoted as $T_D^{(k)}(g)$ for all $ g\in \{1,2,...,m(k)\}$, which accounts for both queueing and transmission delays.

\begin{subequations}\label{eq:2}
In this paper, our objective is to  minimize the average delay by jointly optimizing the subcarrier allocation indicator, $\alpha_{n,k}[t]$, beamforming vectors at the BS, $\mathbf{w}_{n}[t]$, and RIS reflection phase shifts, $\theta_{m}[t]$. Specifically, the problem is formulated as
	
	\begin{align}
		 \min_{{ \{ \theta_m[t], \alpha_{n,k}[t], \mathbf{w}_{n}[t] \} }  } & \frac{1}{K}\sum_{k=1}^{K}\frac{1}{m(k)}\sum_{g=1}^{m(k)}T_D^{(k)}(g)   \notag  \\
		\mathrm{s.t.} \enspace \quad \quad & \quad \alpha_{n,k}[t]\in\{0,1\}, \quad \forall n,k,t,  \label{eq:1B}\\
		& \quad \sum_{k=1}^{K}\alpha_{n,k}[t]\leq1, \quad \forall n,t,  \label{eq:1C}\\
		& \quad \sum_{n=1}^N\|\mathbf{w}_{n}[t]\|^2\leq P_{\max}, \quad \forall t, \label{eq:1D}\\
		& \quad \theta_{m}[t]\in[0, 2\pi], \quad \forall m,t,  \label{eq:1E}
	\end{align}
\end{subequations}
where \eqref{eq:1B} describes the binary constraints of the subcarrier allocation indicaters, \eqref{eq:1C} demonstrates that each subcarrier can be allocated to at most one user, 
\eqref{eq:1D} ensures that the total power consumption of the BS does not exceed its maximum budget $P_{\max}$, while \eqref{eq:1E} limits each RIS element’s reflection phase shift to its allowable range.

\section{DRL-Based Algorithm} \label{section:Deep Reinforcement Learning Based Algorithm}

This section introduces a DRL-based framework for addressing the problem in \eqref{eq:2}, highlighting the core design principles,  corresponding training methodology and the computational complexity of the proposed algorithm.

\begin{figure*}[!t]
	\centering
	\includegraphics[width=0.95\textwidth]{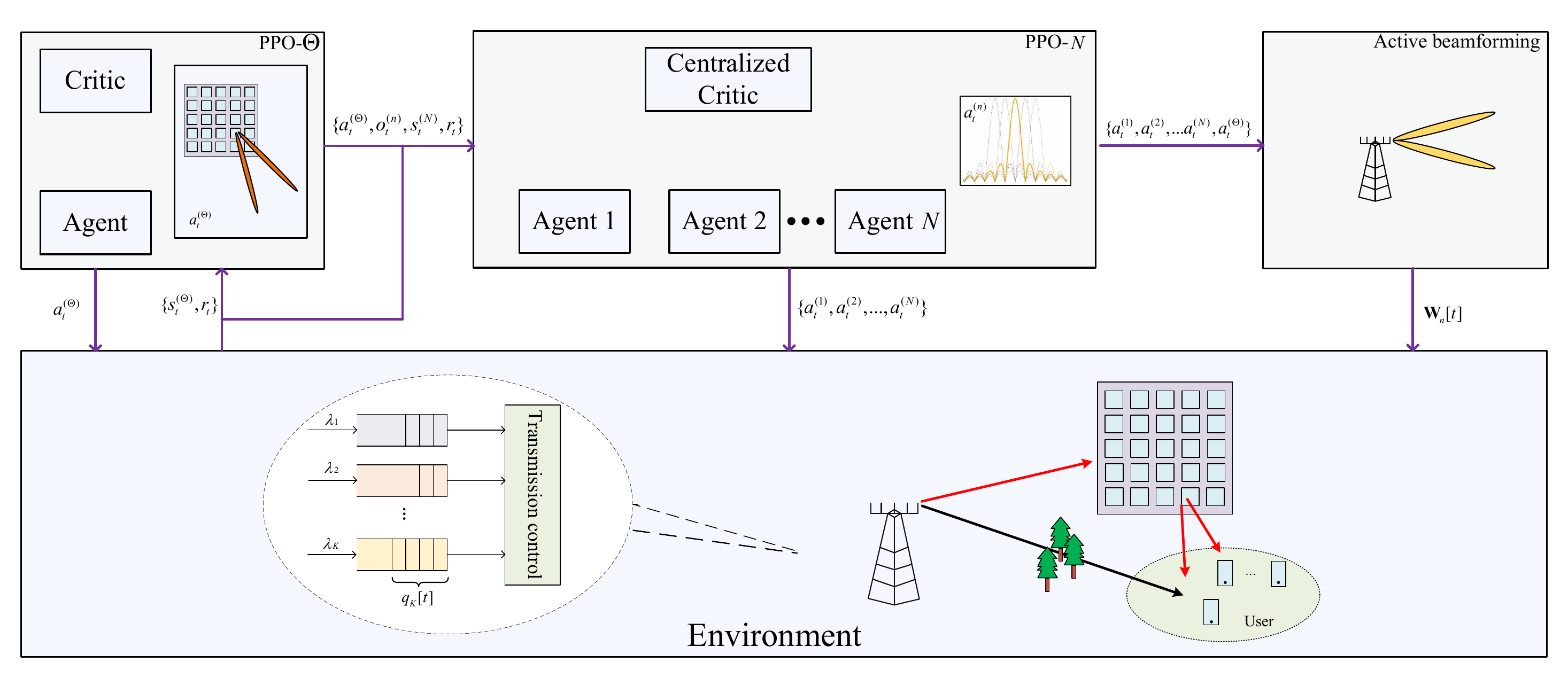}
	\caption{Structure of the proposed DRL framework.}
	\label{Fig:ppo}
\end{figure*}

\subsection{Overall Framework}

The optimization problem presented in equation \eqref{eq:2} necessitates determining the RIS phase shifts and resource allocation for each time slot, where decisions made in the current slot would directly influence the number of packets that can be delivered, thereby affecting the queue length in the subsequent slot. Therefore, problem \eqref{eq:2} is a sequential decision making issue, which can be typically characterized by MDP. To this end, a hybrid RL approach is adopted, wherein a central controller collects information regarding the CSI and the states of all buffers. The central controller, deployed at the BS, functions as an agent responsible for controlling the RIS and subcarrier allocation. It explores the unknown communication environment to acquire experience, which is subsequently utilized to guide the policy design and optimize the RIS reflection phase shifts and subcarrier allocation strategies based on observations of the environment state.  
Specifically, the PPO-$\Theta$ is utilized to predict the RIS reflection phase shifts, while PPO-$N$ is applied to determine the subcarrier allocation indicators. The detailed algorithm framework is illustrated in \Cref{Fig:ppo}. 


\begin{algorithm}[!t]
	\caption{DRL-based Resource Allocation Algorithm}
	\label{alg:1}
	\renewcommand{\algorithmicrequire}{\textbf{Input:}}
	\renewcommand{\algorithmicensure}{\textbf{Output:}}
	\begin{algorithmic}[1]
		
		\REQUIRE  $\bar{\mathbf{h}}^{(d)}_{n,k}[t]$, $\bar{\mathbf{H}}^{(r)}_{n,k}[t]$, $q_k\left[t\right]$, and $\ell_k\left[t\right]$.  
		\ENSURE $\theta_m[t]$, $\alpha_{n,k}[t]$ and $\mathbf{w}_{n}[t]$.    
		
		\STATE Initialize network parameters $\theta^{(\Theta)}$, $\phi^{(\Theta)}$, $\theta^{(N)}$, $\phi^{(N)}$.
		\FOR{each episode}
		\STATE Initialize buffers.
		\STATE Collect $\bar{\mathbf{h}}^{(d)}_{n,k}[t]$, $\bar{\mathbf{H}}^{(r)}_{n,k}[t]$, $q_k\left[t\right]$, and $\ell_k\left[t\right]$.
		\STATE Packets arrive at buffers.
		\FOR{time step $t \in [1, T_{\text{sum}}]$}
		\STATE Obtain action $\theta_m[t]$ from PPO-$\Theta$.
		\STATE Calculate the equivalent channel \( \mathbf{h}_{n,k}^\text{eff} \).
		\STATE Obtain action $\alpha_{n,k}[t]$ from PPO-$N$.
		\STATE Calculate ${\mathbf{w}_{n}[t]}$ according to \Cref{active}.
		\STATE Transmit packets.
		\STATE Packets arrive at buffers.
		\STATE Update buffers of each user.
		\STATE Calculate the reward and obtain new state.
		\STATE Save the experience to the replay buffer 1 and replay buffer 2.
		
		\ENDFOR
		\FOR{agent update times $n \in [1, K_\text{update}]$}
		
		\STATE \makebox[\linewidth][l]{Sample data from replay buffer 1 and update PPO-$\Theta$.}
		
		\STATE \makebox[\linewidth][l]{Sample data from replay buffer 2 and update PPO-$N$.}
		
		\ENDFOR
		\ENDFOR
	\end{algorithmic}
\end{algorithm}

\subsection{PPO-$\Theta$ Design}

The PPO-\(\Theta\) model is employed to optimize the RIS reflection phase shifts. Given the continuous nature of RIS phase shifts, the PPO algorithm is utilized to train an agent comprising an actor network for action selection and a critic network for state value estimation. Specifically, the actor network determines the RIS reflection phase shifts, while the critic network evaluates the effectiveness of the selected policy. In the following, we introduce the state, action, and reward, which are the three fundamental elements for algorithm design.

In wireless communication, CSI significantly impacts the communication performance, making it essential for RIS reflection phase shift design and subcarrier allocation. Moreover, the buffer state information, i.e., the number of queueing packets of each user, and factors that may influcence the buffer state, including the number of current arrival packets, need to be observed in every time slot to make sequential delay-aware decisions. Therefore, the environment state \( s_t^{(\Theta)} \) is defined as the CSI at time slot $t$, including the direct link channel  $\bar{\mathbf{h}}^{(d)}_{n,k}[t]$ and cascaded reflection channel $\bar{\mathbf{H}}^{(r)}_{n,k}[t]$, the number of packets backlogged in each user’s buffer $q_k\left[t\right]$, and the number of current arrival packets $\ell_k\left[t\right]$.

In RL, an action refers to the decision or operation made by an agent in response to a given state of the environment. In the proposed framework, the action of PPO-$\Theta$, denoted as $a_t^{(\Theta)}$, is defined as the RIS reflection phase shift vector
$
\boldsymbol{\theta}[t] = \{\theta_{1}[t], \theta_{2}[t],\cdots , \theta_{M}[t]\}
$.

The reward function plays a pivotal role in DRL, as it directly steers the agent’s learning behavior toward the desired optimization objective. To effectively minimize the average delay in the system, it is essential to align the reward function with the optimization goal.
To this end, the proposed framework defines the reward at each time step as the negative total number of packets queued in all users' buffers, formulated as 

\begin{equation}
	r_t = - \sum_{k=1}^{K} q_k[t].  \label{reward}
\end{equation}
This metric effectively captures the instantaneous queuing state of the network and provides an indirect yet meaningful proxy for overall transmission delay. From a packet-level perspective, if a packet is not transmitted upon arrival, it accumulates delay for each additional time slot as long as it remains in the buffer. Thus, the summation of queue length over all time steps reflects the aggregate queuing delay in the system.
By maximizing the expected return under this reward formulation, the agent is naturally driven to reduce buffer backlogs and equivalently minimize the average delay of all users \cite{fanghao}.

\begin{figure*}[!t]
	\centering
	
	\begin{minipage}[b]{0.9\textwidth}
		\centering
		\subfloat[Actor and critic network architectures of PPO-$\Theta$. While the figure illustrates the structurally similar components of the actor and critic networks in a unified manner for clarity, the two networks are implemented with separate parameters]{\includegraphics[width=\textwidth]{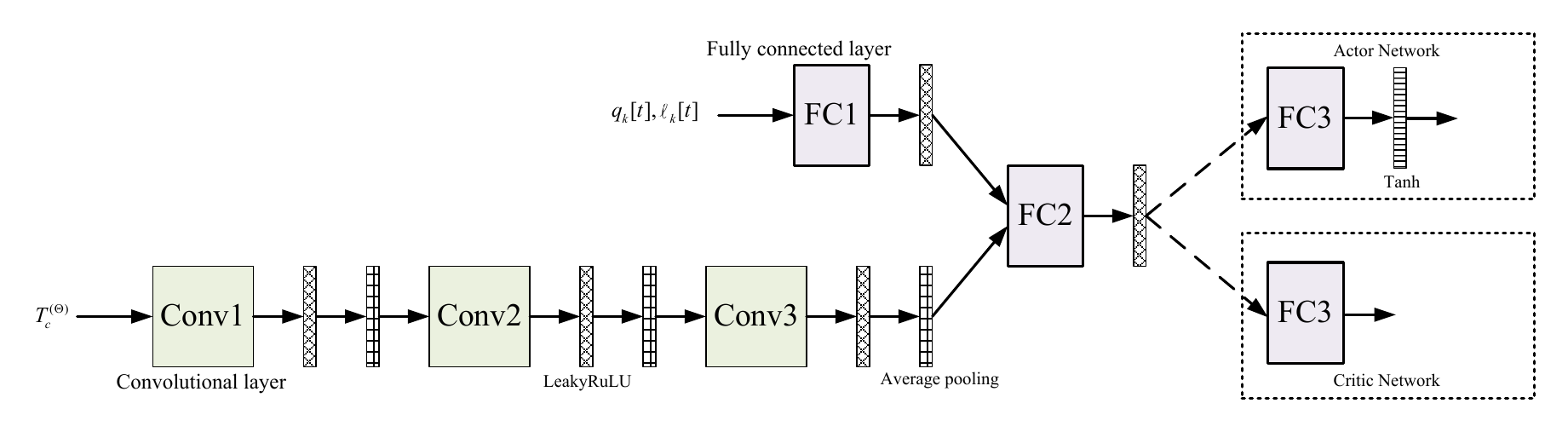} \label{PPOtheta}}
		
	\end{minipage}
	\hspace{0.05\textwidth} 
	
	\begin{minipage}[b]{0.9\textwidth}
		\centering
		\subfloat[Actor network architecture of PPO-$N$]{\includegraphics[width=0.5\textwidth]{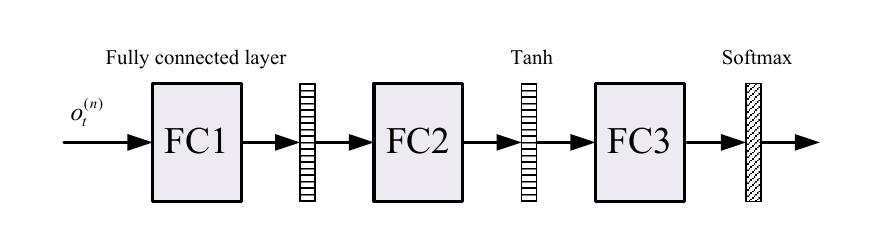}%
			\label{PPOnactor}}
		\hfil
		\subfloat[Centralized critic network architecture of PPO-$N$]{\includegraphics[width=0.5\textwidth]{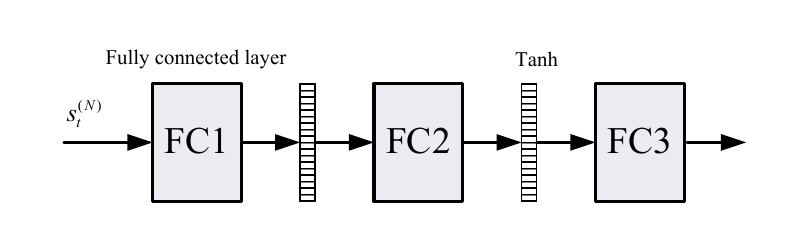}%
			\label{PPONcritic}}
		
	\end{minipage}

	\caption{Network architecture.}
	\label{ppo}
\end{figure*}

At the same time, to ensure effective policy learning and value estimation, the actor and critic networks in PPO-$\Theta$ are carefully designed with tailored neural architectures, as illstrated in Fig. \ref{ppo}\subref{PPOtheta}.

\textbf{\textit{a) Actor network design}}: The input to the actor network corresponds to the state observation $s_t^{(\Theta)}$, which includes both the direct channel $\bar{\mathbf{h}}^{(d)}_{n,k}[t]$ and the cascaded channel $\bar{\mathbf{H}}^{(r)}_{n,k}[t]$, with dimensions of $K \times N \times N_t$ and $K \times N \times M \times N_t$, respectively. Since neural networks cannot directly process complex-valued inputs, each complex channel element is decomposed into its real and imaginary parts.
The resulting tensors are reshaped and combined to construct a $2K$-channel input tensor $T_c^{(\Theta)}$ of size $N N_t \times (M + 1)$, where the channels correspond to the real and imaginary components of the direct and cascaded channels for each user. This tensor is then processed by a convolutional neural network (CNN) to extract spatial features that characterize the wireless environment.
The high-level channel representation extracted by the CNN is subsequently concatenated with the queue length $q_k[t]$ and the transmission power level $\ell_k[t]$, forming a one-dimensional feature vector. This vector is fed into a series of fully connected (FC) layers, which output the parameters of the action probability distribution. Finally, the RIS reflection phase shift action is sampled from this distribution.

\textbf{\textit{b) Critic network design}}: The critic network adopts the same input structure and feature extraction architecture as those of the actor network. However, it differs in its final layer, which is tailored to output a scalar value estimating the state value $V_{\phi^{(\Theta)}} \left(s_{t}^{(\Theta)}\right) \in \mathbb{R}$.

The policy objective function in PPO-$\Theta$ is employed to optimize the actor network and is defined as

\begin{equation}
	\begin{aligned}
		\mathcal{J} (\theta^{(\Theta)}) = \; \mathbb{E}_t \Bigg\{& \min \bigg( 
		 \bar{r}_t^{(\Theta)} (\theta^{(\Theta)}) \hat{A}_t^{(\Theta)}, \\
		& \text{clip} (\bar{r}_t^{(\Theta)} (\theta^{(\Theta)}), 1 - \epsilon^{(\Theta)}, 1 + \epsilon^{(\Theta)}) \hat{A}_t^{(\Theta)} 
		\bigg) \Bigg\},
	\end{aligned}
\end{equation}
where $\bar{r}_t^{(\Theta)} (\theta^{(\Theta)}) = \frac{\pi_{\theta^{\left(\Theta \right)}} \left(a_t^{(\Theta)} | s_t^{(\Theta)}\right)}{\pi_{\theta^{(\Theta)}_{\text{old}}} \left(a_t^{(\Theta)} | s_t^{(\Theta)}\right)}$ denotes the probability ratio between the new and old policies, $\epsilon^{(\Theta)}$ is a clipping factor used in PPO to stabilize learning by preventing excessively large changes of policy update, and $\hat{A}_t^{(\Theta)}$ is the generalized advantage estimation (GAE). The GAE can be computed by

\begin{equation}
\hat{A}_t^{(\Theta)} = \sum_{l=0}^{\infty} (\gamma \lambda)^l \delta_{t+l}^{(\Theta)},
\end{equation}
where $\gamma$ is the discount factor, $\lambda$ is the GAE parameter, while the temporal difference (TD) error $\delta_t^{(\Theta)}$ is defined by

\begin{equation}
\delta_t^{(\Theta)} = r_t + \gamma V_{\phi^{(\Theta)}} \left(s_{t+1}^{(\Theta)} \right) - V_{\phi^{(\Theta)}} \left(s_t^{(\Theta)}\right).
\end{equation}

The critic network, parameterized by $\phi^{(\Theta)}$, is optimized by minimizing the following mean squared error loss

\begin{equation}
\mathcal{L} (\phi^{(\Theta)}) = \mathbb{E}_t \left\{ \bigg( r_t + \gamma V_{\phi^{(\Theta)}} \left(s_{t+1}^{(\Theta)}\right) - V_{\phi^{(\Theta)}} \left(s_t^{(\Theta)}\right) \bigg)^2 \right\},
\end{equation}
where $V_{\phi^{(\Theta)}} \left(s_{t+1}^{(\Theta)}\right)$ and $V_{\phi^{(\Theta)}} \left(s_t^{(\Theta)}\right)$ represent the value estimations of the next and current states, respectively.

\subsection{PPO-$N$ Design}
The PPO-$N$ framework optimizes discrete subcarrier allocation indicators using a discrete PPO algorithm. A single-agent approach faces a huge action space of size \(K^N\) when allocating \(N\) subcarriers to \(K\) users, causing the curse of dimensionality. To address this, a multi-agent approach is used, where each of the \(N\) agents controls one subcarrier’s allocation with an action space of size \(K\), greatly reducing complexity. This design also naturally satisfies constraints \eqref{eq:1B} and \eqref{eq:1C}.

Next, we present the definitions of the global state used by the critic network, the observation state observed by the agents, the reward, and the action.

The global state $s_t^{(N)}$ used by the centralized critic includes the complete equivalent channel state information $\mathbf{h}_{n,k}^\text{eff}[t]$ for all users and subcarriers, along with the global buffer state $q_k[t]$ and packet arrivals $\ell_k[t]$, enabling joint policy evaluation across agents. To reduce the input dimensionality, $\mathbf{h}_{n,k}^\text{eff}[t]$ is defined as $\mathbf{\bar{h}}^{(d)}_{n,k}[t] + \boldsymbol {\mathbf{\varphi}}[t]^T \mathbf{\bar{H}}^{(r)}_{n,k}[t]$. The local observation $o_t^{(n)}$ of agent $n$ comprises the equivalent channel state from the BS to users on subcarrier $n$, buffer length $q_k[t]$, and packet arrivals $\ell_k[t]$.

Since the objective is to minimize the average delay by jointly optimizing the RIS phase shifts and subcarrier allocation, PPO-$N$ and PPO-$\Theta$ share the reward function defined in \eqref{reward}. Since \( N \) agents are used to allocate \( N \) subcarriers, and each subcarrier has \( K \) possible allocation options, the action space $a_{t}^{(n)}$ of each agent is defined as \( \{1, 2, \dots, K\} \).

At the same time, to ensure effective policy learning and value estimation, the actor and critic networks in PPO-$N$ are carefully designed with tailored neural architectures, as described below.

\textbf{\textit{a) Actor network design}}: The input to the actor network corresponds to each observation space \( o_t^{(n)} \). This input is fed into a series of fully connected layers to generate the parameters of the action probability distribution, from which the action is sampled accordingly. The detailed architecture of the actor network is illustrated in Fig. \ref{ppo}\subref{PPOnactor}.

\textbf{\textit{b) Centralized critic network design}}: The input to the centralized critic network corresponds to the global state space \( s_t^{(N)} \). This input is fed into a fully connected neural network to estimate the value of the global state. The critic network outputs a scalar value \( V_{\phi^{(N)}} (s_t^{(N)}) \), which approximates the expected return under the current policy. The detailed architecture of the critic network is illustrated in Fig. \ref{ppo}\subref{PPONcritic}.

In the discrete action space setting, the policy loss function is modified accordingly
\begin{equation}
	\begin{aligned}
		\mathcal{J} (\theta^{(n)}) = & \; \mathbb{E} \Bigg\{ \min \bigg( \bar{r}_{t} \left(\theta^{(n)}\right) \hat{A}_t^{(n)}, \\
		& \quad \text{clip} \left[\bar{r}_{t}^{(n)} \left(\theta^{(n)}\right), 1 - \epsilon^{(n)}, 1 + \epsilon^{(n)}\right] \hat{A}_t^{(n)} \bigg) \Bigg\},
	\end{aligned}
\end{equation}
where $\epsilon^{(n)}$ is a clipping factor, and $\bar{r}_{t}^{(n)} (\theta^{(n)})$ is the probability ratio for agent $n$, given by
\begin{equation}
	\bar{r}_{t}^{(n)} (\theta^{(n)}) = \frac{\pi_{\theta^{(n)}} \left(a_{t}^{(n)} | o_{t}^{(n)}\right)}{\pi_{\theta_{\text{old}}^{(n)}} \left(a_{t}^{(n)} | o_{t}^{(n)}\right)}.
\end{equation}
Note that $\hat{A}_t^{(n)}$ is the GAE for agent $n$, which is computed as
\begin{equation}
	\hat{A}_t^{(n)} = \sum_{l=0}^{\infty} (\gamma \lambda)^l \delta_{t+l}^{(n)},
\end{equation}
where $\gamma$ is the discount factor, $\lambda$ is the GAE parameter, and the TD error is defined by
\begin{equation}
	\delta_t^{(n)} = r_t + \gamma V_{\phi^{(N)}} \left(s_{t+1}^{(N)}\right) - V_{\phi^{(N)}} \left(s_t^{(N)}\right).
\end{equation}

The value function for each agent is optimized by minimizing the following mean squared error loss
\begin{equation}
	\mathcal{L} (\phi^{(N)}) = \mathbb{E}_t \bigg\{  \bigg( r_{t}^{(n)} + V_{\phi^{(N)}} \left(s_{t+1}^{(N)}\right) - V_{\phi^{(N)}} \left(s_t^{(N)}\right) \bigg)^2 \bigg\},
\end{equation}
where $V_{\phi^{(N)}}\left(s_{t + 1}^{(N)}\right)$ and $V_{\phi^{(N)}}\left(s_{t}^{(N)}\right)$ represent the value estimations of the next and current global states, respectively, and $r_{t}^{(n)}$ denotes the reward obtained by agent $n$ from subcarrier allocation decisions.

\begin{figure}[t]
	\centering
	\includegraphics[width=0.5\textwidth]{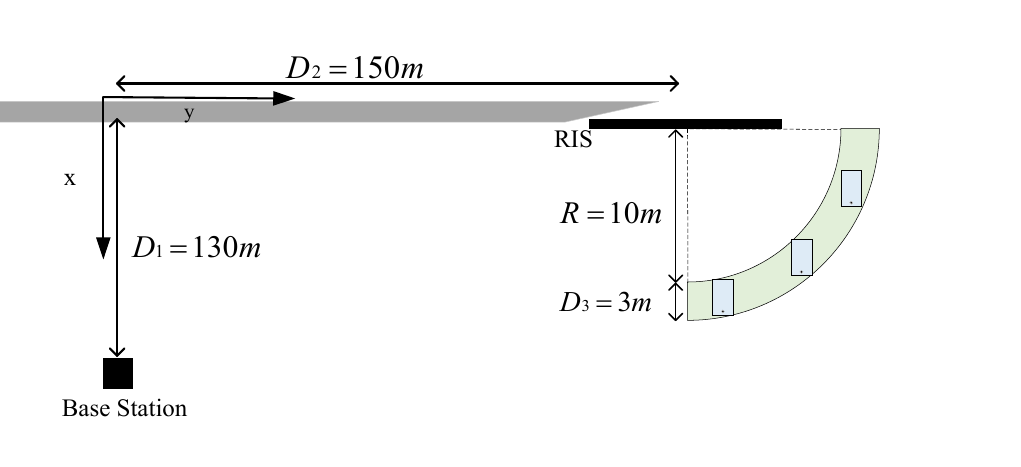}
	\caption{Illustration of the simulated scenario.}
	\label{Fig:position}
\end{figure}

\subsection{Active Beamforming Design}  \label{active}

With the RIS reflection shift and subcarrier allocation indicators produced by PPO-$\Theta$ and PPO-$N$, the BS beamforming vectors ${\mathbf{w}_{n}[t]}$ for subcarrier $n$ at the  time slot $t$ can be designed by leveraging the maximum ratio transmission (MRT) principle in conjunction with the water-filling algorithm, i.e., 

\begin{equation}
	{\mathbf{w}_{n}[t]}=\sqrt{p_{n}[t]}\bar{\mathbf{w}}_{n}[t],
\end{equation}
where $\bar{\mathbf{w}}_{n,q}$ represents the optimal unit-norm beamforming vector and $p_n[t]$ refers to  the transmission power on subcarrier $n$ at time slot $t$. In particular, $\bar{\mathbf{w}}_{n,q}$ can be expressed as,
\begin{equation}
	\bar{\mathbf{w}}_{n}[t]=\frac{\left(\bar{\mathbf{h}}^{(d)}_{n,k}[t]+\boldsymbol{\varphi}[t]^T\bar{\mathbf{H}}^{(r)}_{n,k}[t]\right)^H}{\|(\bar{\mathbf{h}}^{(d)}_{n,k}[t]+\boldsymbol{\varphi}[t]^T\bar{\mathbf{H}}^{(r)}_{n,k}[t])\|},\forall n,t .
\end{equation}
The term $p_{n}[t]$ can be expressed by the following equation, 
\begin{equation}
	p_{n}[t]=\left(\frac{1}{\tau[t]}-\frac{1}{c_{n}[t]}\right)^+, \forall n,t,
\end{equation} 
where $c_{n}[t] = \frac{\|(\bar{\mathbf{h}}^{(d)}_{n,k}[t]+\boldsymbol{\varphi}[t]^T\bar{\mathbf{H}}^{(r)}_{n,k}[t])\|^2}{\sigma^2}$, while $\tau[t]$ is the Lagrange multiplier satisfying,
\begin{equation}
	\begin{aligned}
		\sum_{n=1}^N\left(\frac{1}{\tau[t]}-\frac{1}{c_{n}[t]}\right)^+=P_{\max}, \forall t. 
	\end{aligned}
\end{equation}

\subsection{Training}  \label{train}
Since the RIS phases  have not been optimized in the early stages of training, the system experiences low data rates and overall instability. As a result, the reward value decreases as the time step increases within an episode, making training difficult. To address this, this paper proposes a novel transfer reinforcement learning strategy. Specifically, in the initial training phase, the reward is set to $\min R_k[t]$, while the training process still follows \textbf{\Cref{alg:1}}. The trained PPO-\(\Theta\) model parameters are then saved and reloaded into  \textbf{\Cref{alg:1}} as the new PPO-\(\Theta\) model initialization parameters. The reward is then switched to \eqref{reward}, and training continues until convergence through continuous iterative optimization.

\subsection{Computational Complexity}

The computational complexity of PPO-$\Theta$ is $\mathcal{O}(K N N_t (2M + 1))$, while that of PPO-$N$ is $\mathcal{O}(K N N_t)$. On the other hand, the active beamforming operation incurs a complexity of $\mathcal{O}(N M N_t + N)$. As a result, the overall computational complexity of the proposed algorithm is $\mathcal{O}(K N N_t (2M + 1) + K N N_t + N M N_t + N)$.
Since the max minimum rate benchmark shares the same overall algorithmic structure as the proposed method, its computational complexity is also $\mathcal{O}(K N N_t (2M + 1) + K N N_t + N M N_t + N)$. Moreover, according to \cite{sub} and \cite{sumrate}, the complexity of the max sum rate benchmark is $\mathcal{O}((MN)^{4.5}(KN_t)^{1.5} + (K+1)^4+(K+1)^2KN_tN + NMN_t + N)$.

\begin{table}[t!]
	\centering
	\caption{Hyper-Parameters}
	\label{tab:hyperparams}
	\begin{tabular}{|l|l|}
		\hline
		\textbf{Parameter} & \textbf{Value} \\ \hline
		Number of training episodes & 300 \\ \hline
		PPO-$\Theta$ replay buffer size & 1,000 \\ \hline
		PPO-N replay buffer size & 1,000 \\ \hline
		PPO-$\Theta$ batch size & 64 \\ \hline
		PPO-N batch size & 64 \\ \hline
		Sample reuse  $K_\text{update}$ & 10 \\ \hline
		PPO-$\Theta$ and PPO-$N$ discount factor & 0.9 \\ \hline
		PPO-$\Theta$ and PPO-$N$ GAE parameter & 0.95 \\ \hline
		PPO-$\Theta$ and PPO-$N$ clip parameter & 0.2 \\ \hline
		PPO-$\Theta$ and PPO-$N$ entropy indicater & 0.01 \\ \hline
		Learning rate of all actor networks & 3 \(\times \text{10}^\text{-5} \) \\ \hline
		Learning rate of all critic networks &  5 \( \times \text{10}^\text{-5} \) \\ \hline
		
	\end{tabular}
\end{table}

\section{Numerical Results}   \label{section:Numerical Results}

In this section, a comprehensive evaluation of the performance of the proposed DRL-based algorithm is conducted. Various simulation scenarios and baseline comparisons are considered to demonstrate its effectiveness in optimizing the system performance.  
\subsection{Simulation Settings} \label{set}
In this simulation, as illustrated in \Cref{Fig:position}, a downlink MISO-OFDM system with three single-antenna users is considered, where the users are located in an annular region with inner and outer radii of 10 and 13 meters, respectively, within a 90-degree sector. The vertical and horizontal distances between the BS and the RIS are set to \(D_1=150\) m and \(D_2=130\) m, respectively. The number of reflecting elements on the RIS is \(M=\) 64, and the BS is equipped with 4 antennas. Furthermore, we assume a noise power spectral density of $-174 \ \text{dBm/Hz}$, a transmit power $P_{\text{max}}$ of 10 dBm, and a subcarrier bandwidth of $180 \ \text{kHz}$. The distance between two adjacent antennas on the BS and between adjacent reflecting elements on the RIS is set to half a wavelength. The delay taps are set to $L_0 = 4$, $L_1 = 2$, and $L_2 = 3$, respectively. Additionally, for the direct BS-user link, Rayleigh fading is assumed, while Ricean fading is considered for the reflection channels from the BS to RIS and from RIS to users. The first tap of each channel is set as the line-of-sight (LoS) path, and the remaining taps are non-line-of-sight (NLoS) paths. The Ricean factors for the BS-RIS link and RIS-user link are respectively represented by

\begin{equation}
	{k}_{\mathrm{BR}}=\frac{P_{\mathrm{LoS,BR}}}{P_{\mathrm{NLoS,BR}}},\quad{k}_{\mathrm{RU}}=\frac{P_{\mathrm{LoS,RU}}}{P_{\mathrm{NLoS,RU}}},
\end{equation}

\noindent where, \(P_{\text{LoS, BR}}\), \(P_{\text{NLoS, BR}}\), \(P_{\text{LoS, RU}}\), and \(P_{\text{NLoS, RU}}\) represent the power of the LoS and NLoS paths for the respective links. In this simulation, \({k}_{\mathrm{BR}}=4\) dB and \({k}_{\mathrm{RU}}=5\) dB are assumed. The large-scale fading is defined as \( \beta = \beta_0 \left(\frac{d}{d_0}\right)^{-\xi} \), where \( \beta_0 = -30 \) dB, \( d_0 = 1 \) m, \( d \) represents the link distance, and \( \xi \) is the path loss exponent. The path loss exponents for the BS-user link, BS-RIS link, and RIS-user link are \( \xi_0 = 3.8 \), \( \xi_1 = 2.2 \), and \( \xi_2 = 2.4 \), respectively.

The packet capacity $L$ is set to 512 bits. In each episode, the communication duration is set to $T_{\text{sum}}=$ 1,000 time slots, with each time slot lasting 1 ms. Since the position of user remains nearly fixed within 1 second, we assume that the user positions are fixed throughout each episode.\footnote{Note that, to accelerate the convergence of the algorithm, the user positions are randomly generated at each step within an episode during the training process.} The packet arrival rate for each user is set to \( \lambda = 9.5 \). Throughout this work, unless stated otherwise, identical arrival rates are assumed for all users, i.e., \( \lambda = \lambda_1 = \lambda_2 = \cdots = \lambda_K \). The detailed hyperparameters are presented in \Cref{tab:hyperparams}.

\begin{figure}[t]
	\centering
	\includegraphics[width=0.45\textwidth]{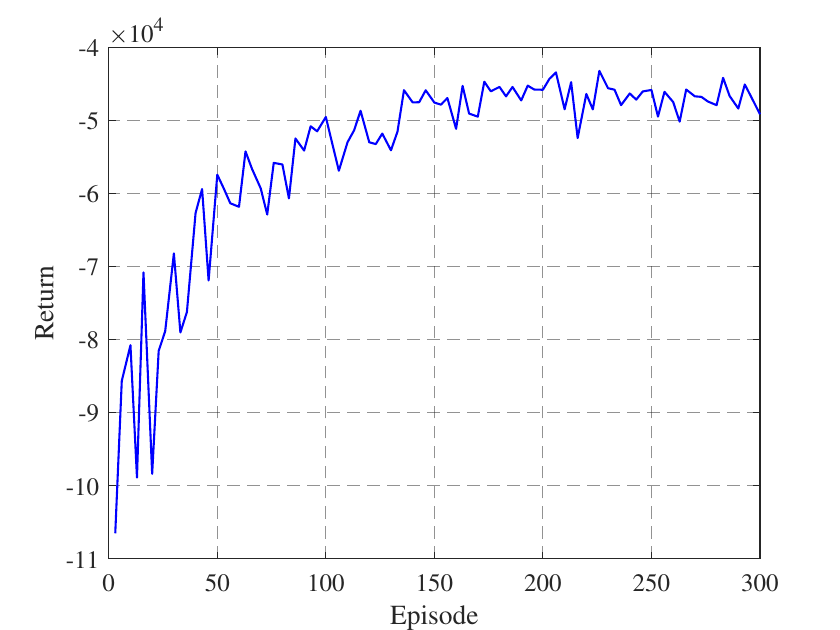}
	\caption{Return for each training episode with an increasing number of iterations.}
	\label{Fig:return}
\end{figure}

\subsection{Baseline Methods}

To reflect the performance of our proposed method, the following baseline methods are considered as follows:

\begin{itemize}
	\item \textbf{Max sum rate}: The algorithm proposed in \cite{sub} and \cite{sumrate} is respectively employed for subcarrier allocation and RIS reflection shift optimization, while MRT is used for beamforming at the BS.
	
	\item \textbf{Max minimum rate}: This algorithm is the hybrid DRL algorithm with the reward to be $\min R_k[t]$, as described in \Cref{train}, and it has not undergone transfer learning.
	
	\item \textbf{Random}: The RIS reflection shift and subcarrier allocation are set randomly. The active beamforming at the BS is designed using the MRT.
	
	\item \textbf{Without RIS}: This scheme represents a system that lacks RIS assistance, i.e., the number of RIS reflecting elements is set to \( M = 0 \).

\end{itemize}

\begin{figure}[t]
	\centering
	\includegraphics[width=0.45\textwidth]{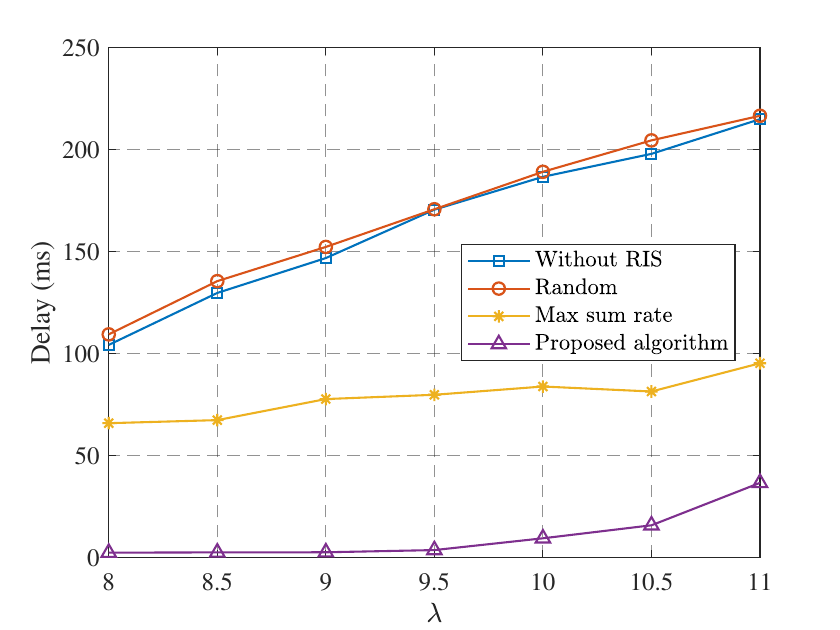}
	\caption{Average delay performance under varying packet arrival rates $\lambda$.}
	\label{Fig:delay}
\end{figure}

\begin{figure}[t]
	\centering
	\includegraphics[width=0.45\textwidth]{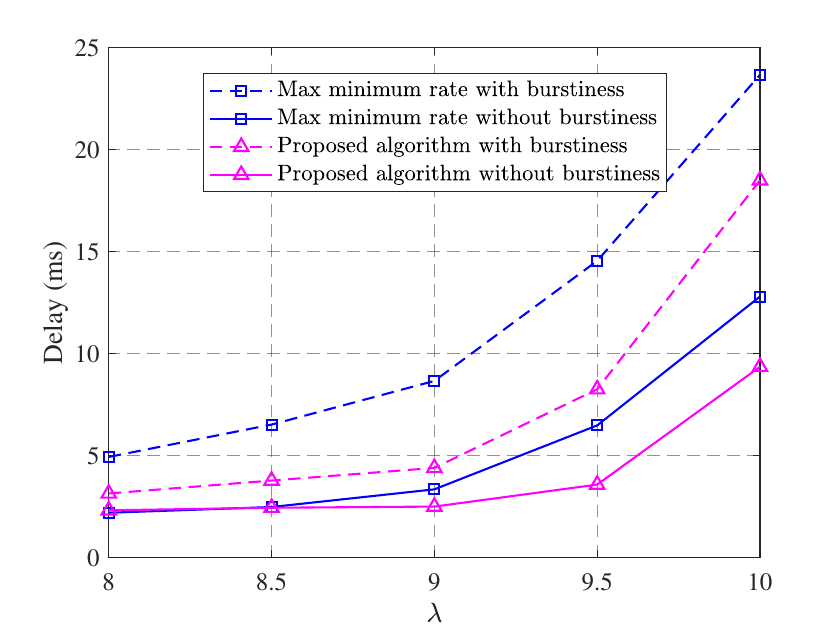}
	\caption{Average delay performance under varying packet arrival rates $\lambda$.}
	\label{Fig:delayEmergency}
\end{figure}

\subsection{Convergence and Validity Analysis}

To validate the effectiveness and convergence of the proposed method, experiments are conducted.  \autoref{Fig:return} shows the cumulative reward for each training episode. The figure demonstrates a progressive increase in the cumulative reward throughout the training process, indicating the effectiveness of our proposed method. Besides, the performance gradually stabilizes in around 150 training episodes, despite some fluctuations due to channel variations and changes in the number of packets.

\subsection{Latency  Performance Analysis}

\autoref{Fig:delay} illustrates the average delay performance under varying packet arrival rates $\lambda$. The proposed algorithm consistently achieves the lowest average delay, demonstrating its superior capability in minimizing latency. In contrast, the ``without RIS" scheme results in the highest delay, emphasizing the critical role of RIS in improving the delay performance. Although the ``max sum rate" method performs better than baseline schemes, such as ``random" and ``without RIS", its overall delay performance is clearly inferior to that of the proposed method. This is because it prioritizes the maximization of system throughput without considering user fairness or queue stability, which leads to higher delays for certain users. The ``random" method yields persistently high delays, further underlining the importance of intelligent and delay-aware resource allocation.
Moreover, as  $\lambda$ increases, the number of packets arriving at the buffer within each time slot grows accordingly, leading to aggravated buffer backlog and, consequently, resulting in an incremental increase in the average delay.

\begin{figure*}[!t]
	\centering
	
	\begin{minipage}[b]{\textwidth}
		\centering
		\subfloat[The backlogged packets in each buffer of the proposed method]{\includegraphics[width=0.5\textwidth]{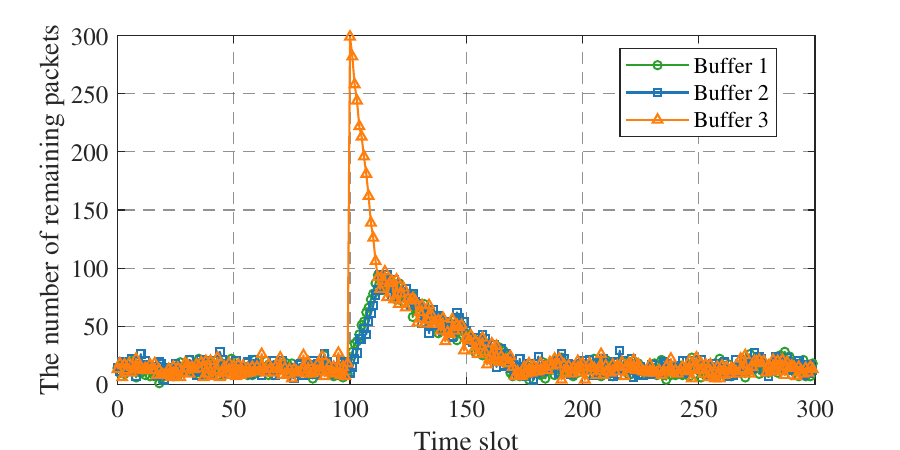}%
			\label{Fig:packagePropsed}}
		\hfil
		\subfloat[The backlogged packets in each buffer of the max minimum rate method]{\includegraphics[width=0.5\textwidth]{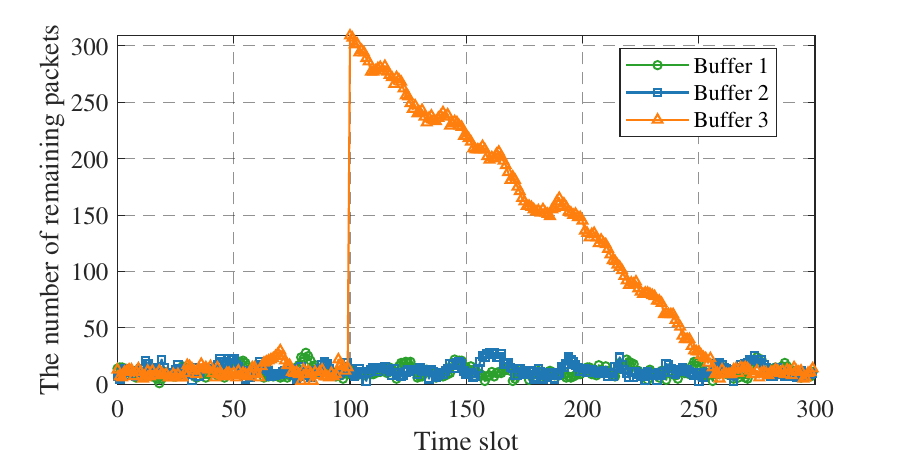}%
			\label{Fig:packageMaxMin}}
		
	\end{minipage}
	\hspace{0.05\textwidth} 
	
	\hspace{0.05\textwidth} 
	
	\begin{minipage}[b]{\textwidth}
		\centering
		\subfloat[Transmission rates of the proposed method]{\includegraphics[width=0.5\textwidth]{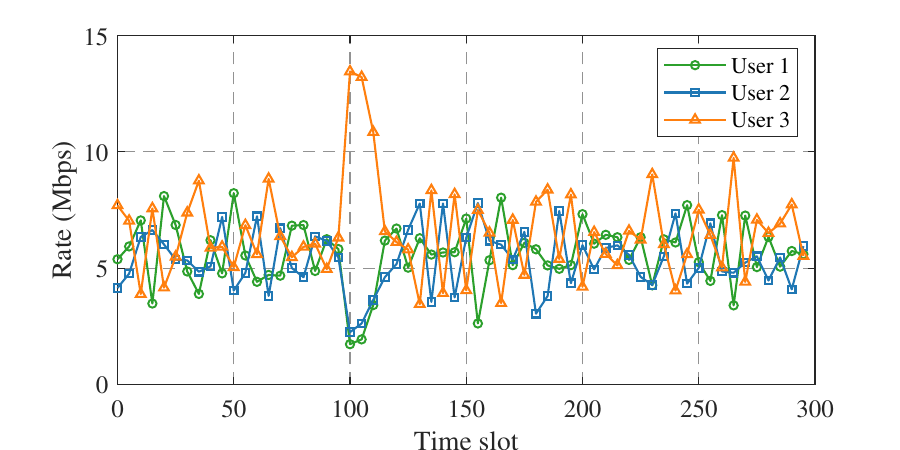}%
			\label{rateProposed}}
		\hfil
		\subfloat[Transmission rates of the max minimum rate method]{\includegraphics[width=0.5\textwidth]{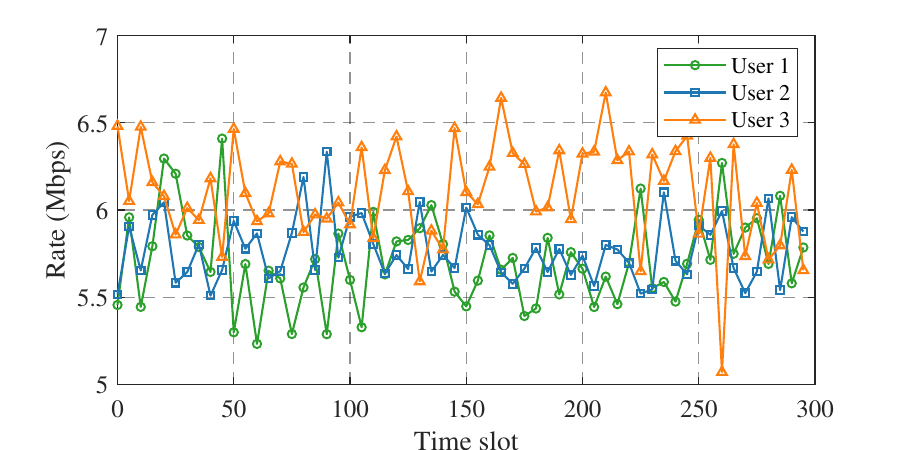}%
			\label{rateMaxMin}}
		
	\end{minipage}
	
	\caption{The change of the backlogged packets and transmission rates in each buffer over time steps of the proposed method and max minimum rate method in brustiness scenarios.}
	\label{fig:emergrncy}
\end{figure*}

\begin{figure*}[!t]
	\centering
	
	\begin{minipage}[b]{\textwidth}
		\centering
		\subfloat[The backlogged packets in each buffer of the proposed method]{\includegraphics[width=0.5\textwidth]{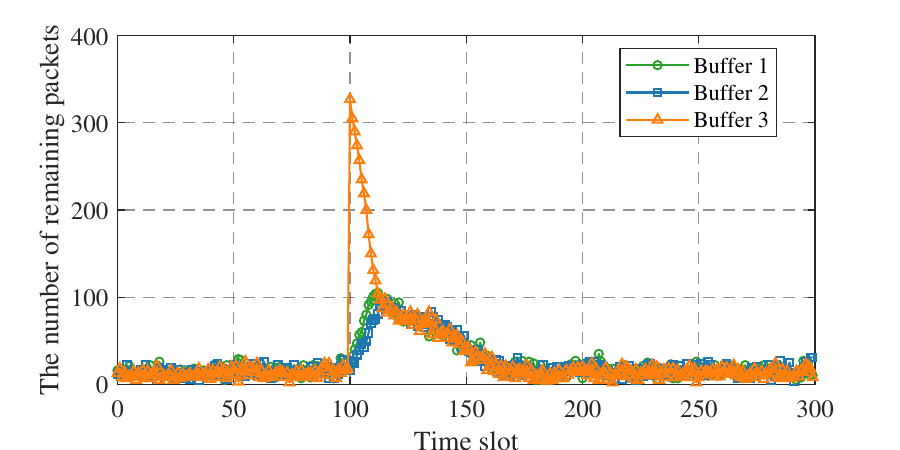}%
			\label{Fig:learningrate}}
		\hfil
		\subfloat[The backlogged packets in each buffer of the max minimum rate method]{\includegraphics[width=0.5\textwidth]{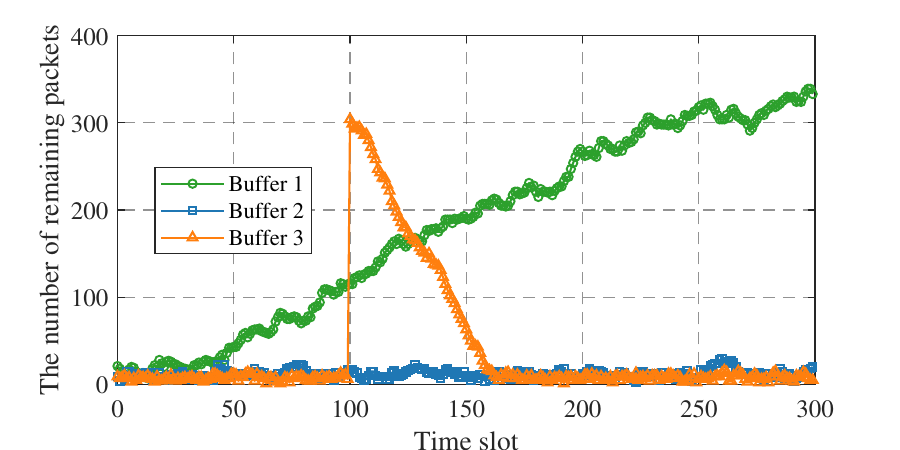}%
			\label{Fig:learningratevs}}
		
	\end{minipage}
	\hspace{0.05\textwidth} 
	
	\hspace{0.05\textwidth} 
	
	\begin{minipage}[b]{\textwidth}
		\centering
		\subfloat[Transmission rates of the proposed method]{\includegraphics[width=0.5\textwidth]{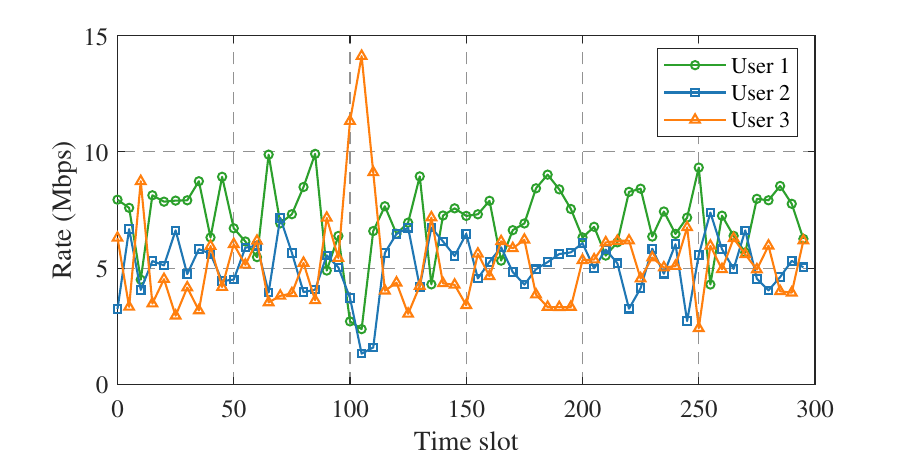}%
			\label{SE-Res}}
		\hfil
		\subfloat[Transmission rates of the max minimum rate method]{\includegraphics[width=0.5\textwidth]{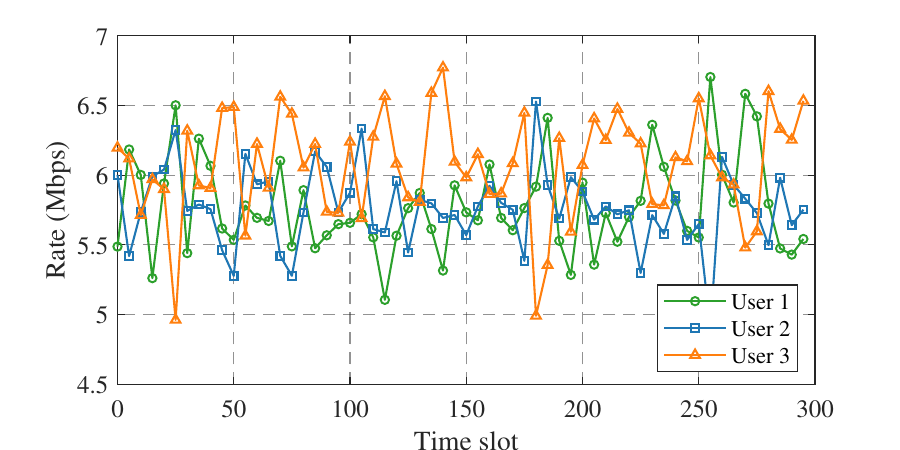}%
			\label{Symbols}}
		
	\end{minipage}
	
	\caption{The change of the backlogged packets and transmission rates in each buffer over time steps of the proposed method and max minimum rate method in different traffic flows scenarios.}
	\label{fig:main}
\end{figure*}

\begin{figure}[t]
	\centering
	\includegraphics[width=0.45\textwidth]{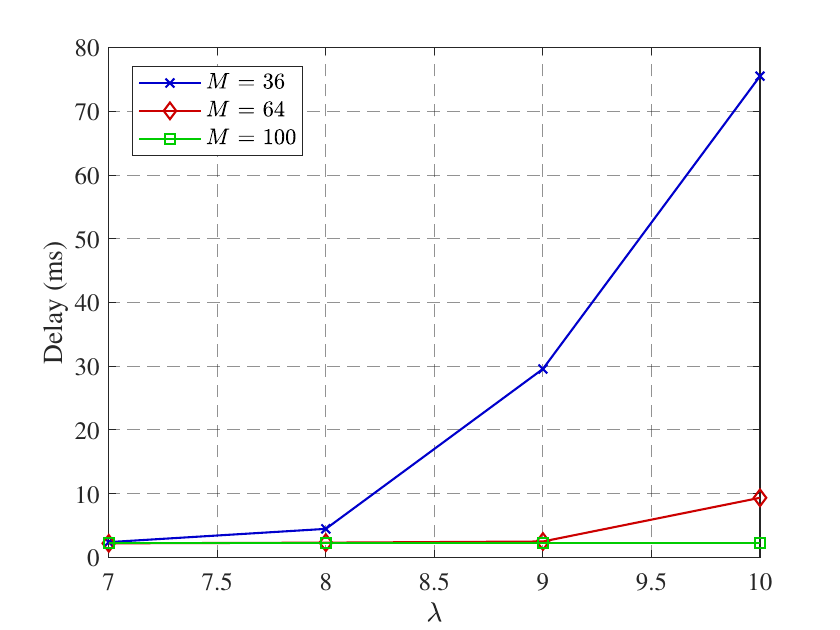}
	\caption{Impact of the number of reflection elements $M$ on the delay performance.}
	\label{Fig:element}
\end{figure}

\subsection{Latency Analysis in Brustiness Scenarios}

To evaluate the proposed algorithm under bursty traffic, we simulate sudden data arrivals by injecting 300 packets into buffers 1, 2, and 3 at time slots 100, 400, and 700, respectively. \autoref{Fig:delayEmergency} presents a comparison of the delay performance between the proposed algorithm and the ``max minimum rate" algorithm under different scenarios. It can be observed that, regardless of whether burstiness is present, the proposed algorithm consistently demonstrates superior performance, effectively validating its  effectiveness  in handling burstiness.

To further explain this advantage, we analyze a specific case by examining buffer backlogs and transmission rates during the first 300 time slots. As shown in \Cref{fig:emergrncy}\subref{Fig:packagePropsed}, the proposed algorithm effectively reduces backlog by dynamically adjusting resource allocation in response to bursty arrivals. In contrast, \Cref{fig:emergrncy}\subref{Fig:packageMaxMin} shows persistent congestion in buffer 3 under the ``max minimum rate'' approach, indicating inefficient allocation.
\autoref{fig:emergrncy}\subref{rateProposed} further shows that the proposed method prioritizes users with heavier backlogs, rapidly increasing their transmission rates, such as user 3, leading to quick backlog clearance. Conversely, \Cref{fig:emergrncy}\subref{rateMaxMin} reveals that the baseline allocates resources more rigidly, ignoring buffer states and failing to relieve congestion effectively, which results in higher delay.

\subsection{Impact of Reflection Elements $M$}

\autoref{Fig:element} illustrates the impact of the number of RIS reflection elements on the average delay performance under varying arrival rates \(\lambda\). As observed, when \(M = 36\), the average delay increases sharply once \(\lambda\) exceeds 9.0, suggesting that the system struggles to accommodate high traffic loads. With \(M = 64\), the delay grows more moderately but still experiences a considerable rise at elevated arrival rates. In contrast, when \(M = 100\), the delay remains consistently low across all values of \(\lambda\), demonstrating that \(M = 100\) is sufficient enough to support the considered service load with \(\lambda\) less than 10.
This improvement is primarily attributed to the enhanced passive beamforming gain resulting from the increased number of RIS reflection elements, thereby improving the effective data rate. As a result, the system is able to accommodate higher traffic loads while maintaining low-latency communication performance.

\subsection{Latency Analysis under Different Traffic Flows}

\begin{figure}[t]
	\centering
	\includegraphics[width=0.45\textwidth]{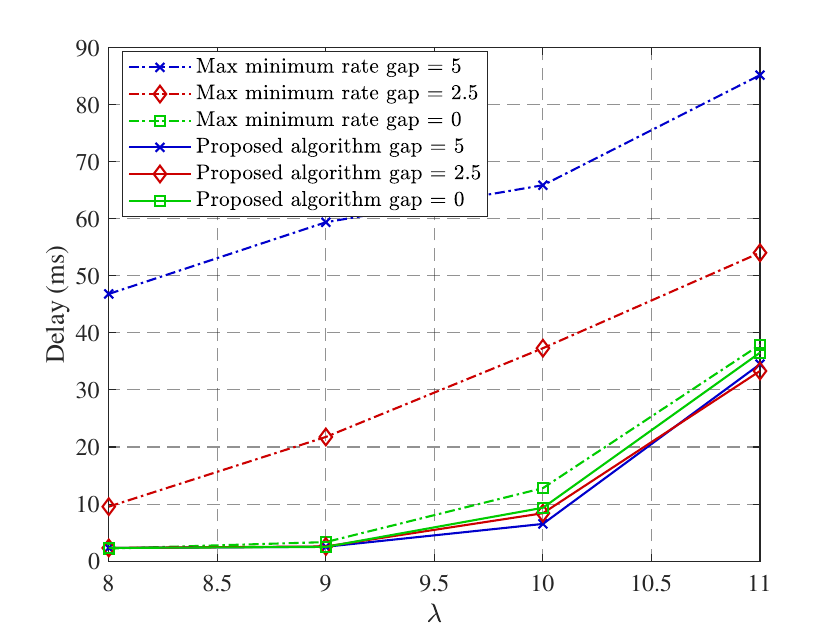}
	\caption{Average delay performance under varying packet arrival rates $\lambda$.}
	\label{Fig:gap}
\end{figure}

To evaluate the performance of the proposed algorithm under varying user arrival rates, we set the packet arrival rates \(\lambda_1 = \lambda + \text{gap}\), \(\lambda_2 = \lambda\), and \(\lambda_3 = \lambda - \text{gap}\). As shown in \Cref{Fig:gap}, the delay of the ``max minimum rate'' method increases with larger gaps, while the proposed algorithm maintains consistently low delay, indicating strong robustness even when trained under uniform arrival rates.

To further explain this advantage, we analyze a specific case by examining buffer backlogs and transmission rates during the first 300 time slots.
In \Cref{fig:main}\subref{Fig:packagePropsed}, the proposed algorithm keeps buffer backlogs balanced despite differing arrival rates, showing its fairness. Upon the burst at time slot 100, the algorithm quickly reallocates resources to clear buffer 3. In contrast,  \Cref{fig:main}\subref{Fig:packageMaxMin} shows that the ``max minimum rate'' approach leads to severe congestion in buffer 1 due to static rate-based allocation.
Further, \Cref{fig:main}\subref{rateProposed} reveals that the proposed method allocates more resources to high arrival rate users and dynamically adapts to queue states. Before time slot 100, user 3 has a lower rate due to low arrival, but receives rapid support when the burst occurs. However, \Cref{fig:main}\subref{rateMaxMin} shows that the ``max minimum rate'' method fails to allocate resources and optimize beamforming based on buffer states, leading to increased delays.

\begin{table*}[!t]
	\caption{The robustness of proposed approach in terms of average delay}
	\label{tab1}
	\centering
	\begin{tblr}{
			colspec = {|X[c]|X[c]|X[c]|X[c]|X[c]|},
			hlines, vlines
		}
		Algorithm & Delay taps ($L_0 = 6$, $L_1 = 3$, $L_3 = 4$) & Distribution ($R = 15$ m, $D_3 = 3$ m)  & Ricean factor (\(k_{\mathrm{BR}}=3\) dB, \(k_{\mathrm{RU}}=4\) dB)\\
		Proposed algorithm & 9.6902   ms & 63.3849 ms  & 6.2181 ms\\
		Max sum rate       & 91.1677 ms  & 72.8252 ms  & 78.7498 ms\\ 
		Max minimum rate   & 27.3113 ms  & 64.5808 ms  & 9.4219 ms \\
		Random             & 167.4868. ms & 207.0432 ms & 174.2089 ms\\  
	\end{tblr}
\end{table*}

\subsection{Jitter Analysis}
\begin{figure}[t]
	\centering
	\includegraphics[width=0.45\textwidth]{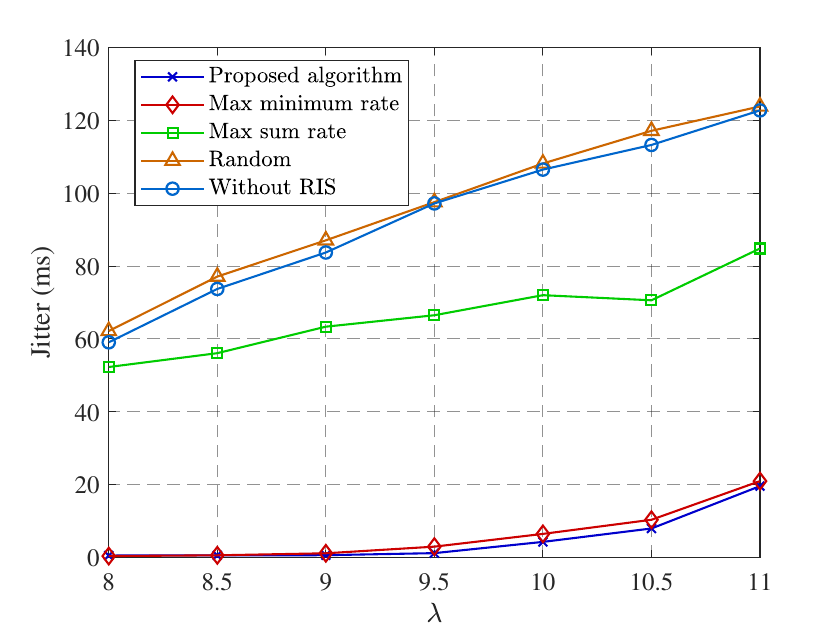}
	\caption{Jitter performance under varying packet arrival rates $\lambda$.}
	\label{Fig:jitter}
\end{figure}

To evaluate the jitter performance of our proposed algorithm, we conduct measurements and analysis based on the jitter definition provided in \cite{jitter}, as given by
\begin{equation}
	\frac{1}{K}\sum_{k=1}^{K}\sqrt{\frac{1}{m(k)}\sum_{g=1}^{m(k)}\bigg(T_D^{(k)}(g) - \frac{1}{m(k)} \sum_{g=1}^{m(k)} T_D^{(k)}(g) \bigg)^2}. \label{jitter}
\end{equation}
As illustrated in \Cref{Fig:jitter}, the proposed algorithm consistently achieves the lowest jitter across all values of $\lambda$, demonstrating its superior capability in balancing resource allocation and maintaining latency stability. In contrast, the ``max-min rate'' scheme yields slightly higher jitter, while the ``max sum rate'' approach, although effective in optimizing the overall throughput, results in significantly increased jitter. The ``random'' allocation and ``without RIS'' schemes exhibit the worst jitter performance, underscoring the effectiveness of both learning-based optimization and RIS assistance in enhancing system delay stability. 
Moreover, as the arrival rate $\lambda$ increases, jitter also exhibits a rising trend, indicating that the system becomes increasingly unstable under higher traffic loads and struggles to maintain consistent delay performance.

\subsection{Robustness Validation}

\Cref{tab1} presents a robustness evaluation of the proposed algorithm in terms of average delay under three distinct transmission scenarios: varying delay taps ($L_0 = 6$, $L_1 = 3$, $L_3 = 4$), spatial distribution parameters ($R = 15\,\text{m}$, $D_3 = 3\,\text{m}$), and Ricean fading factors ($k_{\text{BR}} = 3\,\text{dB}$, $k_{\text{RU}} = 4\,\text{dB}$). The per-user arrival rate is fixed at $\lambda = 9.5$, with all other parameters following the baseline in \Cref{set}.
Across all scenarios, the proposed algorithm consistently achieves the lowest delay, significantly outperforming benchmark methods. In contrast, the ``max sum rate'' approach leads to notably higher delays, while the ``max minimum rate'' shows moderate performance. The ``random'' strategy results in the worst delay across all settings.
These results confirm the strong adaptability and delay minimization capability of the proposed algorithm under diverse transmission conditions. Its robustness against variations in different delay taps, user distributions, and fading characteristics highlights its potential for real-world deployment in dynamic wireless environments.

\section{Conclusion}  \label{section:Conclusion}

This paper introduced a novel DRL-based approach to minimize the average delay while ensuring user fairness in different traffic flows.  Unlike traditional methods, our proposed solution extends the state information to encompass a comprehensive set of factors, including CSI, the number of backlogged packets in buffer and the number of current arrival packets. To address the challenges posed by the hybrid action space, we developed a hybrid DRL framework, where the action space consists of both continuous and discrete components.  Specifically, PPO-$\Theta$ was employed to optimize the continuous RIS reflection phase shifts, while PPO-$N$ was designed to handle the discrete subcarrier allocation decisions.  Furthermore, to tackle the curse of dimensionality arising from multi-dimensional discrete variables, we introduced a multi-agent strategy to optimize subcarrier allocation decisions efficiently. Through extensive numerical evaluations, our method demonstrated significant improvements in reducing the average delay while maintaining user fairness, surpassing existing baseline methods.  Additionally, the results highlighted the superior stability and adaptability of our approach.  This work provides a robust and scalable solution to the delay minimization in RIS-assised OFDM systems, leveraging advanced machine learning techniques to optimize resource allocation in next-generation wireless networks, paving the way for more efficient and equitable network management.

\bibliographystyle{IEEEtran}
\bibliographystyle{unsrt}

\bibliography{references.bib}

\end{document}